\documentclass[conference]{IEEEtran}
\IEEEoverridecommandlockouts
\usepackage{cite}
\usepackage{amsmath,amssymb,amsfonts}
\usepackage{algorithmic}
\usepackage{graphicx}
\usepackage{textcomp}
\usepackage{xcolor}

\usepackage{multirow}
\usepackage{longtable}
\usepackage{rotating}
\usepackage{color}
\usepackage{makecell}
\usepackage{lineno}
\graphicspath{{figures/}}

\def\BibTeX{{\rm B\kern-.05em{\sc i\kern-.025em b}\kern-.08em
    T\kern-.1667em\lower.7ex\hbox{E}\kern-.125emX}}
\begin{document}

\title{Infrared and visible image fusion using Latent Low-Rank Representation
}

\author{\IEEEauthorblockN{Hui Li, Xiao-Jun Wu\textsuperscript{*}}
\IEEEauthorblockA{\textit{Jiangsu Provincial Engineering Laboratory of Pattern Recognition and Computational Intelligence,} \\
\textit{School of IoT engineering, Jiangnan University,}\\
Wuxi, China \\
lihui@stu.jiangnan.edu.cn, wu\_xiaojun@jiangnan.edu.cn}

}

\maketitle

\begin{abstract}
Infrared and visible image fusion is an important problem in the field of image fusion which has been applied widely in many fields. To better preserve the useful information from source images, in this paper, we propose a novel image fusion method based on latent low-rank representation(LatLRR) which is simple and effective. Firstly, the source images are decomposed into low-rank parts(global structure) and salient parts(local structure) by LatLRR. Then, the low-rank parts are fused by weighted-average strategy to preserve more contour information. Then, the salient parts are simply fused by sum strategy which is a efficient operation in this fusion framework. Finally, the fused image is obtained by combining the fused low-rank part and the fused salient part. Compared with other fusion methods experimentally, the proposed method has better fusion performance than state-of-the-art fusion methods in both subjective and objective evaluation. The Code of our fusion method is available at https://github.com/hli1221/imagefusion\_Infrared\_visible\_latlrr.
\end{abstract}


\begin{figure*}[ht]
\centering
\includegraphics[width=\linewidth]{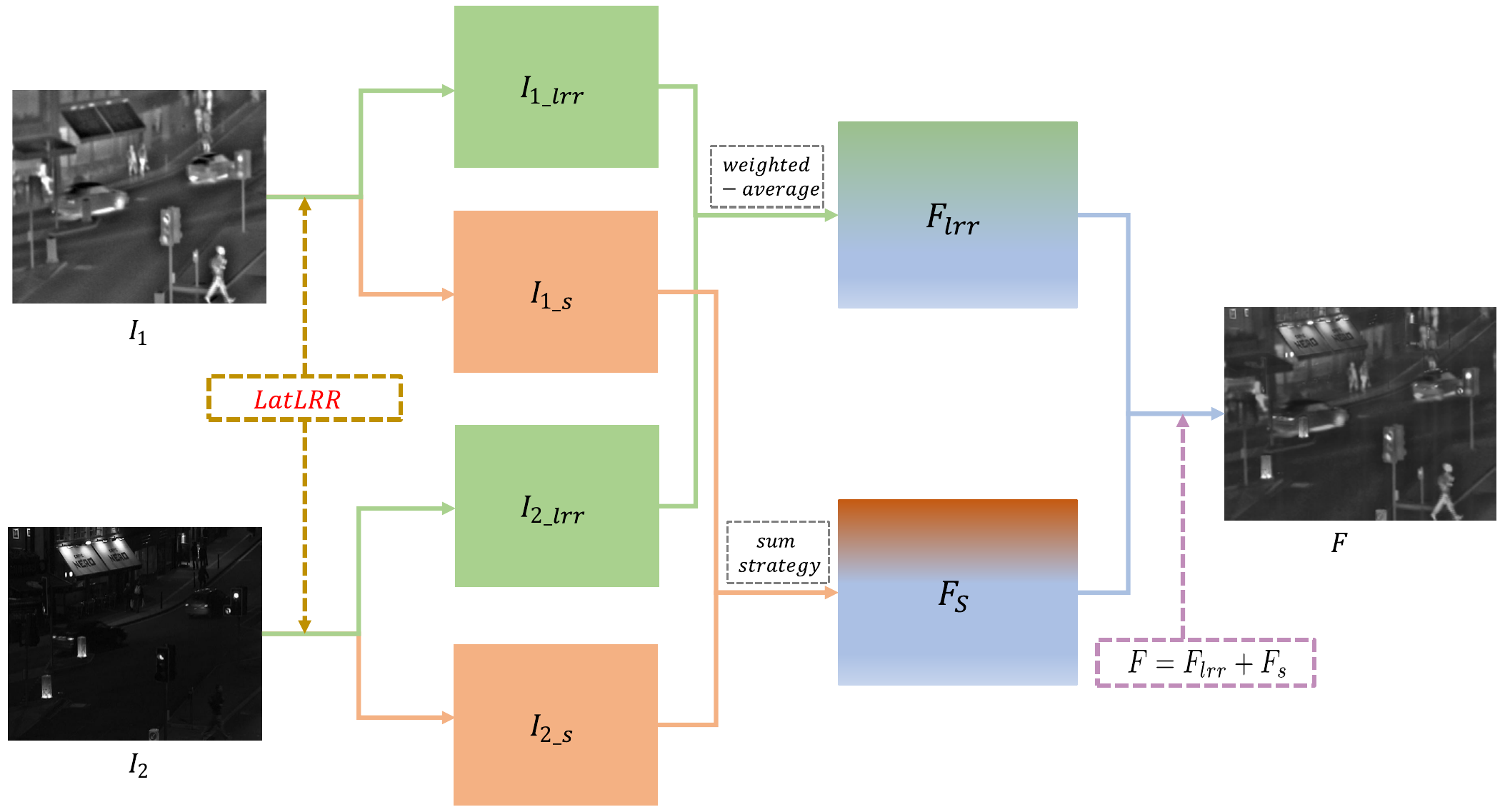}
\caption{The framework of proposed method.}
\label{fig:IVIF_Latlrr}
\end{figure*}

\section{Introduction}

In multi-sensor image fusion field, the infrared and visible image fusion is an important task. It has been widely used in many applications, such as surveillance, object detection and target recognition. The main purpose of image fusion is to generate a single image which contains the complementary information from multiple images of the same scene\cite{Shutao2017}. In infrared and visible image fusion, it is a key problem to extract the saliency object from infrared image and visible image. Many fusion methods have been proposed recently.

As we all know, the most commonly used method in image fusion are multi-scale transforms, such as discrete wavelet transform(DWT)\cite{DWT2005}, contourlet transform\cite{contourlet2010}, shift-invariant shearlet transform\cite{shearlet2014} and quaternion wavelet transform\cite{quaternion wavelet2013} etc. Due to the conventional transform methods has not enough detail preservation ability, Lou et al. \cite{luo2016novel} \cite{nonshearlet2017} proposed a fusion method based on contextual statistical similarity and nonsubsampled shearlet transform which can obtain the local structure information from source image and get good fusion performance. For infrared and visible image fusion, Bavirisetti et al.\cite{twoscale2016} proposed a fusion method based on two-scale decomposed and salient detection, they used mean filter and median filter to extract the base layers and detail layers, and used visual salient to obtain weight maps. Finally, the fused image is obtained by calculating these three parts. Besides, Zhang et al.\cite{morphological2017} proposed a morphological gradient based fusion method. This method used the different morphological gradient operator to obtain the focus region, defocus region and focus boundary region, respectively. Then the fused image is obtained by using an appropriate fusion strategy. Besides, Luo et al.\cite{Luo2017JVCIR} proposed an novel image fusion method based on HOSVD and edge intensity. And a fuzzy transform based fusion method was proposed by Manchanda et al.\cite{Manchanda2018}. These methods both obtain better fusion performance in subjective and objective evaluation \cite{zheng2006nearest}\cite{sun2011quantum}\cite{wang2003initial}.

Recently, with the rise of compressed sensing, image fusion methods based on representation learning has attracted great attention. The most common methods of representation learning are sparse representation(SR). Zong et al.\cite{medicalclassified2017} proposed a novel medical image fusion method based on SR. The Histogram of Oriented Gradient(HOG) features were used to classify the image patches and learn several sub-dictionaries. Then this method used the $l_1$-norm and choose-max strategy to reconstruct fused image. In addition, there are many methods based on combining SR and other tools which are pulse coupled neural network(PCNN)\cite{ivsr2014}, low-rank representation(LRR)\cite{lrr2017} and shearlet transform\cite{ivsidt2016}. In sparse domain, the joint sparse representation \cite{jsrsd2017} and cosparse representation\cite{gao2017} were also applied into image fusion.

Although the SR based fusion methods obtain good fusion performance, these methods still have drawback, such as the ability of capturing global structure is limited. To address this problem, we introduce a new representation learning technique, latent low-rank representation (LatLRR)\cite{latentLrr2011}, to infrared and visible image fusion task. Unlike low-rank representation(LRR)\cite{Lrr2010}, the LatLRR can extract the global structure information and the local structure information from source images.

In deep learning field, deep features of the source images are used to reconstruct the fused image \cite{li2018infrared} \cite{li2018densefuse} \cite{ma2019fusiongan}. In \cite{csr2016}, Yu Liu et al. proposed a fusion method based on convolutional sparse representation(CSR). The CSR is different from deep learning methods, but the features extracted by CSR are still deep features. In addition, Yu Liu et al. \cite{cnn2017} also proposed a convolutional neural network(CNN)-based fusion method. Image patches which contain different blur versions of the input image are used to train the network and get a decision map. The fused image is obtained by using the decision map and the source images. However, these deep learning-based methods still have drawbacks since the network is difficult to train when the training data is not enough, especially for infrared and visible image fusion task.

In this paper, we propose a novel fusion method based on LatLRR in infrared and visible image fusion. The source images are decomposed into low-rank parts(global structure) and salient parts(local structure) by LatLRR. Then we use different fusion strategy to fuse low-rank parts and salient parts. Finally, we use fused low-rank parts and fused salient parts to reconstruct the fused image. The experimental results demonstrate that our proposed method has better fusion performance than other fusion methods.

This paper is structured as follows. In Section 2, we give a brief introduction to LatLRR theory. In Section 3, the proposed LatLRR based image fusion method is introduced in detail. The experimental results are shown in Section 4. Finally, Section 5 draws the conclusions.

\section{Latent Low-Rank Representation Theroy}

In 2010, Liu et al.\cite{Lrr2010} proposed LRR theroy, but this representation method can not preserve the local structure information. So in 2011, the author proposed LatLRR theroy\cite{latentLrr2011}, and the global structure and local structure can be extracted by LatLRR from raw data.

In reference\cite{latentLrr2011}, the LatLRR problem is reduced to solve the following optimization problem,
\begin{eqnarray}\label{equ:lrr}
  	\min_{Z,L,E}||Z||_*+||L||_*+\lambda||E||_{1} \\
    s.t.,X=XZ+LX+E \nonumber
\end{eqnarray}

\noindent where $\lambda>0$ is the balance coefficient, $||\cdot||_*$ denotes the nuclear norm which is the sum of the singular values of matrix and $||\cdot||_1$ is $l_1$-norm. $X$ denotes observed data matrix,$Z$ is the low-rank coefficients, $L$ is the salient coefficients, and $E$ is the sparse noisy part. Eq.(\ref{equ:lrr}) is solved by the inexact Augmented Lagrangian Multiplier (ALM). Then the low-rank part $XZ$ and the salient part $LX$ are obtained by Eq.(\ref{equ:lrr}).

\begin{figure}[ht]
\centering
\includegraphics[width=\linewidth]{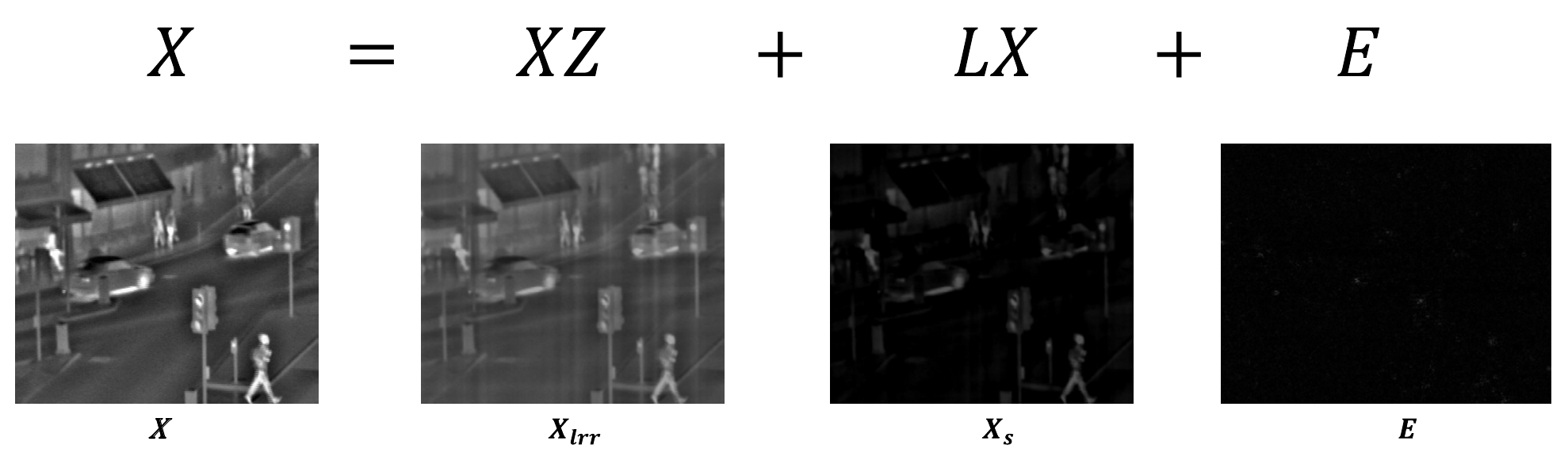}
\caption{The LatLRR decomposed operation.}
\label{fig:LatlrrDecomposed}
\end{figure}

\section{The Proposed Fusion Method}

In this section, the proposed fusion method is presented in detail. And the fusion processing of low-rank parts and salient parts will be introduced in this sections. 

The LatLRR decomposed operation is shown in Fig.\ref{fig:LatlrrDecomposed}. The low-rank part and salient part for input image $X$ are obtained by LatLRR.

As shown in Fig.\ref{fig:LatlrrDecomposed}, $X$ is the input image, $X_{lrr}=XZ$ and $X_s=LX$ denote the low-rank part and salient part which are obtained by LatLRR. When we get the low-rank parts and salient parts from each source image, we will use fusion strategy to fused these two parts, respectively. And the framework of the proposed fusion method is shown in Fig.\ref{fig:IVIF_Latlrr}.

The two source images are denoted as $I_1$(infrared) and $I_2$(visible). Firstly, the low-rank part $I_{C\_lrr}$ and salient part $I_{C\_s}$ for each source image are obtained by LatLRR, where $C\in{\{1,2\}}$. Then the low-rank parts and salient parts are fused by weighted-average strategy, respectively. Finally, the fused image $F$ will be reconstructed by adding the fused low-rank part $F_lrr$ and salient part $F_s$.

\subsection{Fusion of low-rank parts}
Low-rank parts for source images contain more global structure information and brightness information. So, in our fusion method, we use weighted average strategy to obtain the fused low-rank part. The fused low-rank part is calculated by Eq.(\ref{equ:lrrparts_fusion}),
\begin{eqnarray}\label{equ:lrrparts_fusion}
  	F_{lrr}(i,j)=w_1I_{1\_lrr}(i,j)+w_2I_{2\_lrr}(i,j)
\end{eqnarray}

\noindent where $(i,j)$ denotes the corresponding position of the coefficients of $I_{1\_lrr}$, $I_{2\_lrr}$ and $F_{lrr}$, respectively. And $w_1$ and $w_2$ represent the weight values for these coefficients($I_{1\_lrr}$ and $I_{2\_lrr}$). To preserve the global structure and brightness information, and to reduce the redundant information, in this paper, we choose $w_1=0.5$ and $w_2=0.5$.

\subsection{Fusion of salient parts}
\begin{figure}[!ht]
\centering
\includegraphics[width=\linewidth]{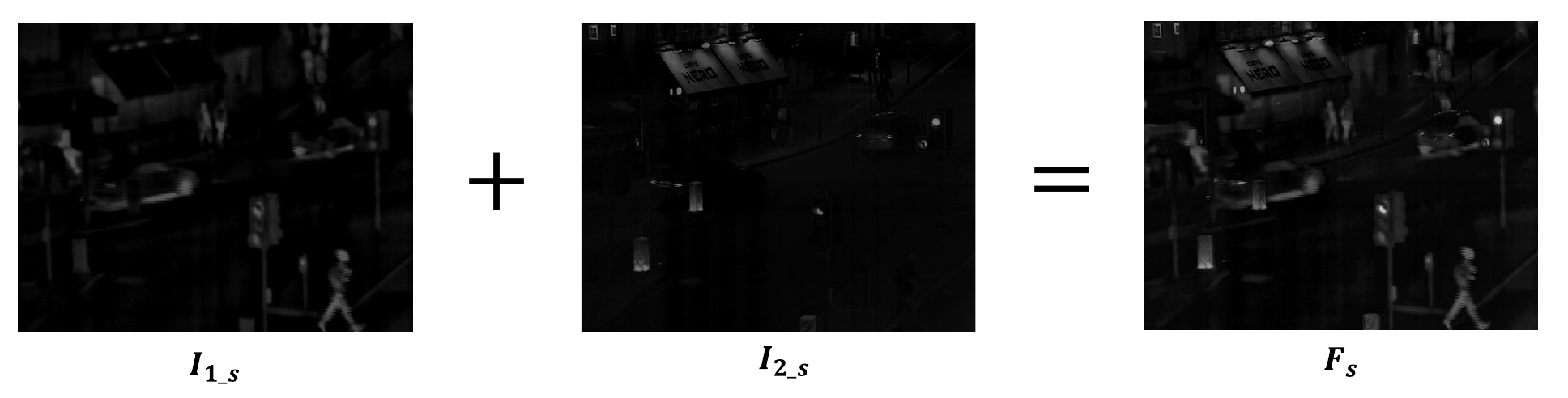}
\caption{The fusion procedure of salient parts.}
\label{fig:saliencyparts_fusion}
\end{figure}

The salient parts contain the local structure information and salient features, as shown in Fig.\ref{fig:saliencyparts_fusion}. 

In Fig.\ref{fig:saliencyparts_fusion}, the salient part $I_{1\_s}$ contains more salient features from $I_1$ and it is the same situation for $I_{2\_s}$ and $I_2$. The salient features from source images are complementary information and need to be contained in fused images without loss. So in Eq.(\ref{equ:saliencyparts_fusion}), we simply use sum strategy to fuse the salient parts.
\begin{eqnarray}\label{equ:saliencyparts_fusion}
  	F_{s}(i,j)=s_1I_{1\_s}(i,j)+s_2I_{2\_s}(i,j)
\end{eqnarray}

In Eq.(\ref{equ:saliencyparts_fusion}), the $(i,j)$ represents the corresponding position of the coefficients of $I_{1\_s}$, $I_{2\_s}$ and $F_s$. And $w_1$ and $w_2$ represent the weight values for coefficients of $I_{1\_s}$ and $I_{2\_s}$, respectively. To preserve more local structure and saliency features in fused images, in this paper, we choose $s_1=1$ and $s_2=1$ in Eq.(\ref{equ:saliencyparts_fusion}). In the next subsection, we will introduce why we use sum strategy and choose $s_1=s_2=1$.

\subsection{The reason of choose sum strategy}
\begin{figure}[!ht]
\centering
\includegraphics[width=\linewidth]{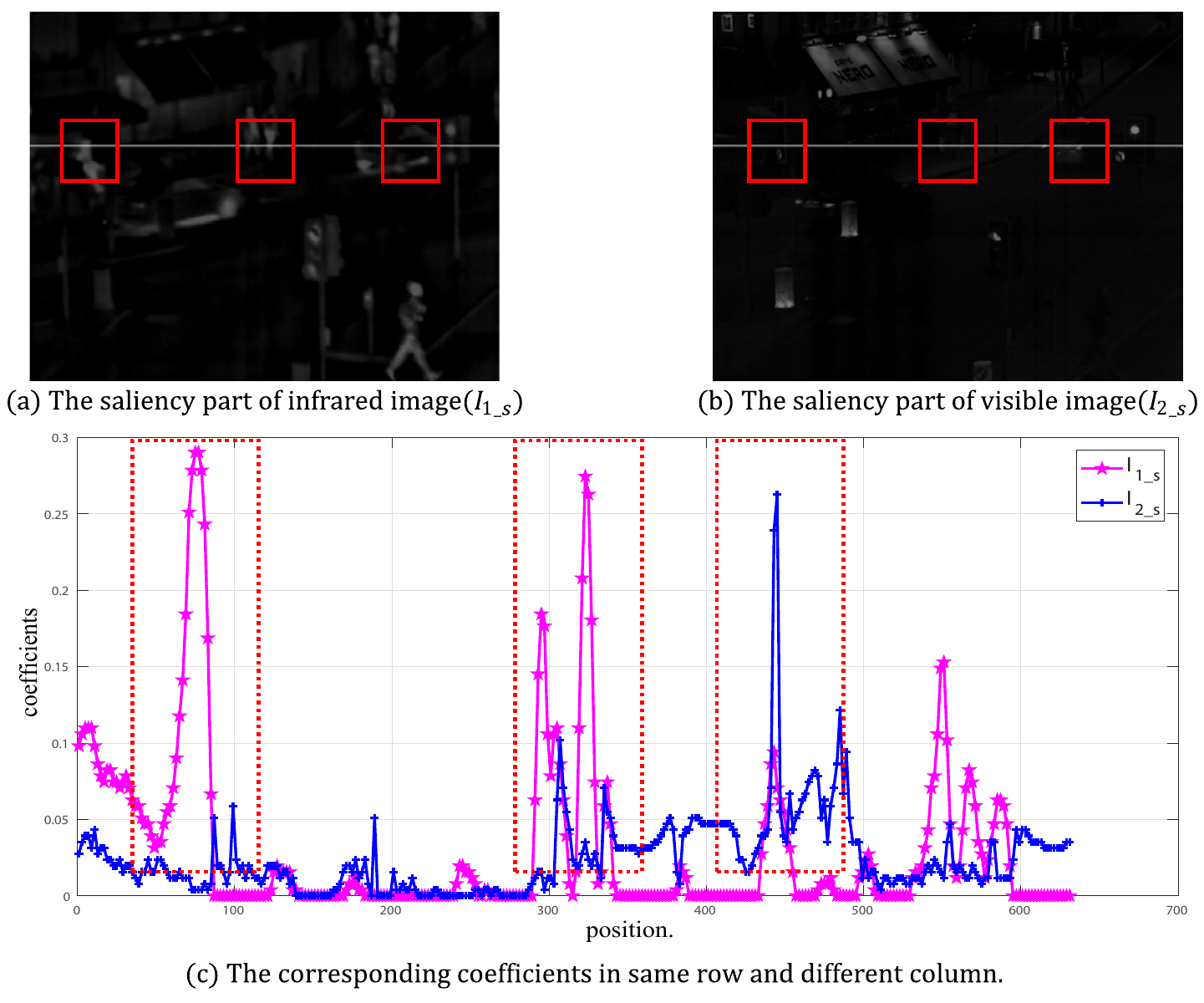}
\caption{Example of coefficients corresponding to the same row in infrared and visible salient parts, (a)$I_{1\_s}$ and (b)$I_{2\_s}$ denote the saliency parts, (c) plot of two saliency parts values in same row.}
\label{fig:spartvalue}
\end{figure}

In this section, we will explain why we simply use the sum strategy to fuse the salient parts. We choose the coefficients from two salient parts in same row, as shown in Fig.\ref{fig:spartvalue}(a) and Fig.\ref{fig:spartvalue}(b). Fig.\ref{fig:spartvalue}(c) indicate the salient part values in the same row and different columns.

In Fig.\ref{fig:spartvalue}(a) and Fig.\ref{fig:spartvalue}(b), the white line denotes the row which we choose in Fig.\ref{fig:spartvalue}(c). And in Fig.\ref{fig:spartvalue}(c), the blue line and orange line indicate the coefficients of infrared salient part($I_{1\_s}$) and visible salient part($I_{2\_s}$), respectively. From Fig.\ref{fig:spartvalue}(a) and Fig.\ref{fig:spartvalue}(b), in the first and second red boxes, the infrared salient part contains more saliency features which means the coefficients of infrared saliency part are larger than visible salient part in corresponding position and the values of visible salient part are very small, as shown in Fig.\ref{fig:spartvalue}(c)(the position in $[50,100]$ and $[270,350]$). In the third red box, the visible salient part contains more salient features, which means the coefficients of visible salient part are larger than infrared salient part in corresponding position and the values of infrared saliency part are very small, as shown in Fig.\ref{fig:spartvalue}(c)(the position in $[400,500]$).

\begin{figure}[!ht]
\centering
\includegraphics[width=\linewidth]{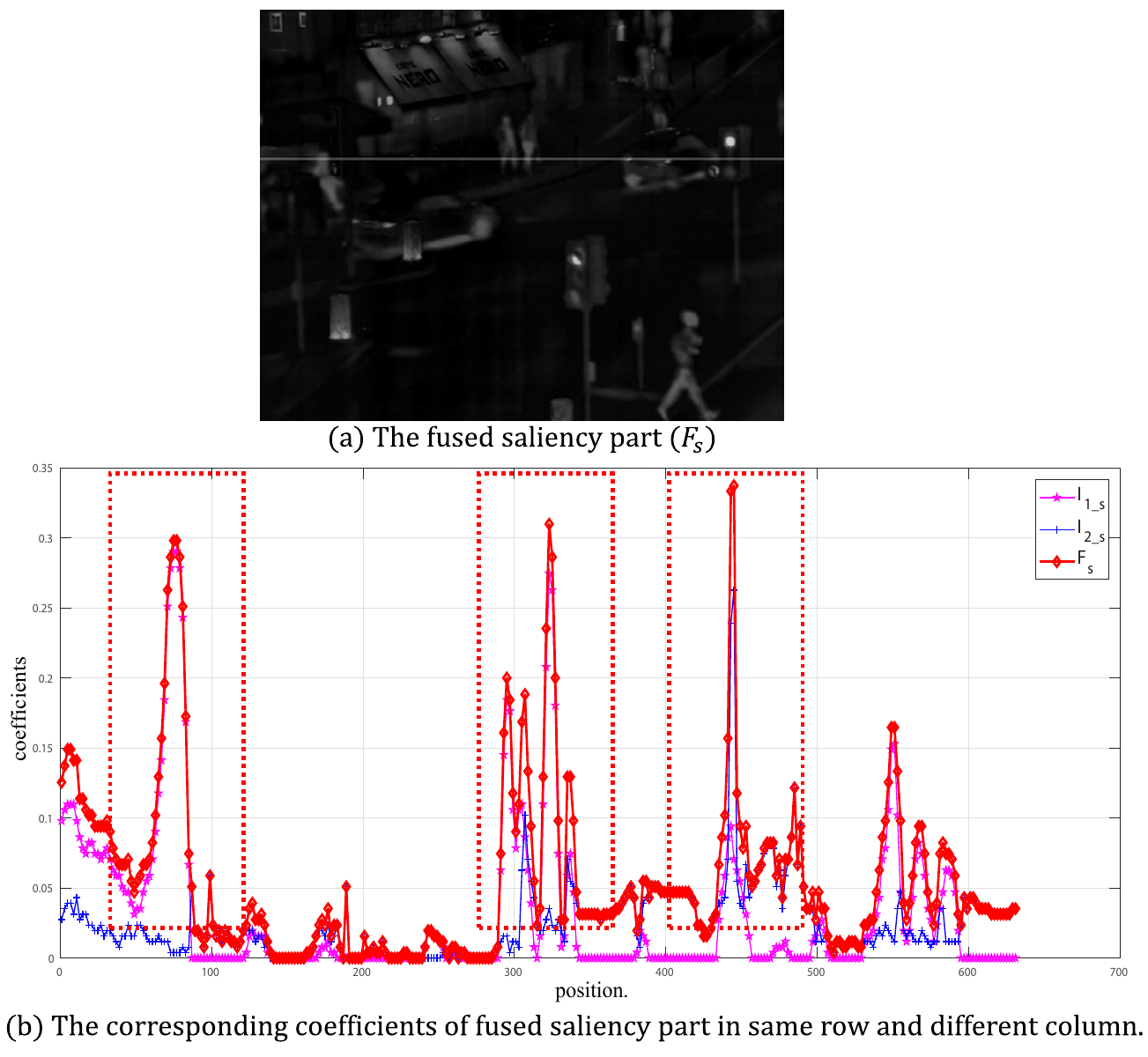}
\caption{Example of coefficients corresponding to the same row in fused salient parts, (a) the fused salient parts, (b) plot of fused saliency parts values in the same row.}
\label{fig:fpartvalue}
\end{figure}

The fused salient part and the corresponding coefficients are shown in Fig.\ref{fig:fpartvalue}. Fig.\ref{fig:fpartvalue}(a) is the fused salient part which uses sum strategy, and the red line in Fig.\ref{fig:fpartvalue}(b) indicates the coefficients of fused salient part in the same row.

As we discussed in subsection 3.2, the salient features from source images need to be contained in the fused image without loss. If we choose weight-average strategy, the salient features will be reduced. And if we choose sum strategy to fuse salient parts, the salient features will be contained in fused salient part without loss. Furthermore, the features will be increased by sum strategy, as shown in Fig.\ref{fig:fpartvalue}(b).

\subsection{Reconstruction}
When the fused low-rank part $F_{lrr}$ and salient part $F_s$ are obtained by Eq.(\ref{equ:lrrparts_fusion}) and Eq.(\ref{equ:saliencyparts_fusion}), the fused image $F$ will be reconstructed by Eq.(\ref{equ:recons}),
\begin{eqnarray}\label{equ:recons}
  	F(i,j)=F_{lrr}(i,j)+F_{s}(i,j)
\end{eqnarray}

\subsection{Summary of the Proposed Fusion Method}
We summarize the proposed fusion method based on LatLRR as follows:

	1) The source images are decomposed by LatLRR to obtain the low-rank parts $I_{C\_lrr}$ and the salient parts $I_{C\_s}$, where $C\in{\{1,2\}}$.
	
	2) We choose weighted-average fusion strategy and sum strategy to fuse low-rank parts and salient parts, respectively. Then the fused low-rank part $F_{lrr}$ and the fused salient part $F_s$ are obtained.
	
	3) Finally, the fused image are obtained by Eq.(\ref{equ:recons}).

\section{Experimental Results}
In this section, first of all, we introduce the detailed experimental settings. Then the subjective and objective methods are adopted to assess the fusion performance. Finally, the experimental results are analyzed visually and quantitatively.

\subsection{Experimental Settings}
\begin{figure}[!ht]
\centering
\includegraphics[width=\linewidth]{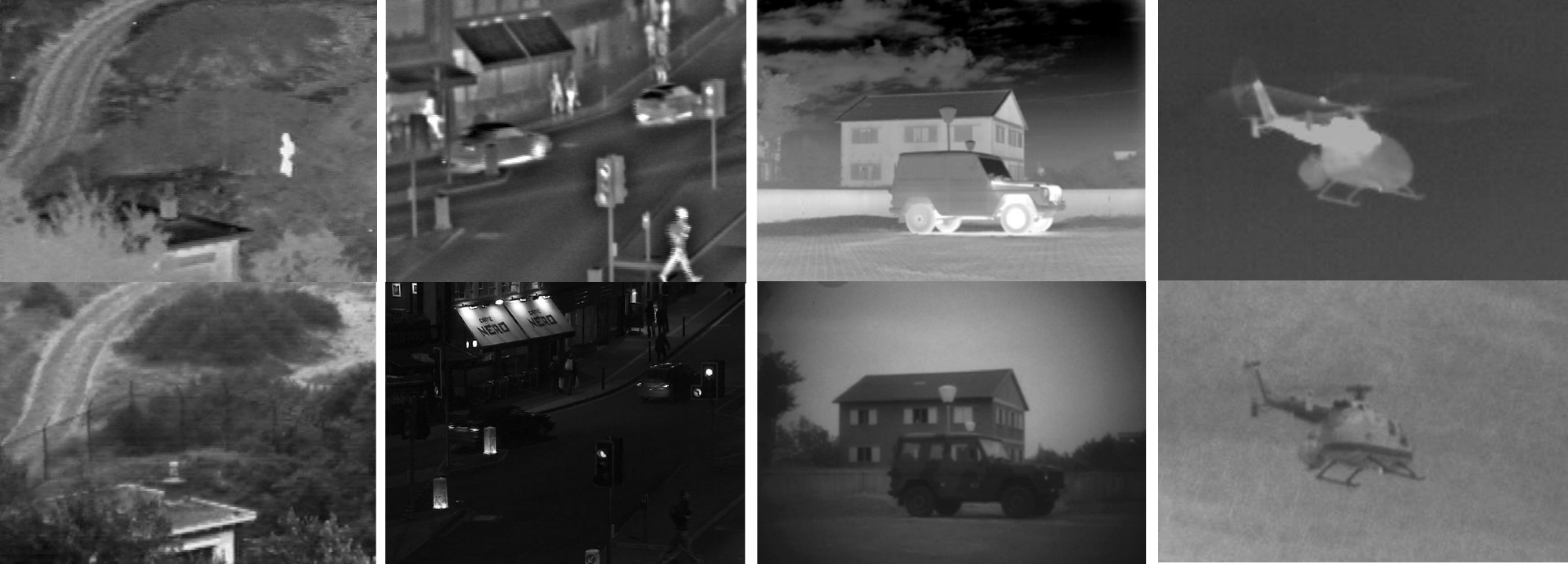}
\caption{Eight pairs of source images from 21 pairs.}
\label{fig:example4}
\end{figure}

Firstly, the source infrared and visible images were collected from \cite{ivwls2017} and \cite{TNO2014}. And the number of our source images are 21 pairs\cite{GithubHuiLi}. Because the number of infrared and visible images are too many to show all of them, so we just take several examples from these images, as shown in Fig.\ref{fig:example4}.

\begin{figure*}[!ht]
\centering
\includegraphics[width=\linewidth]{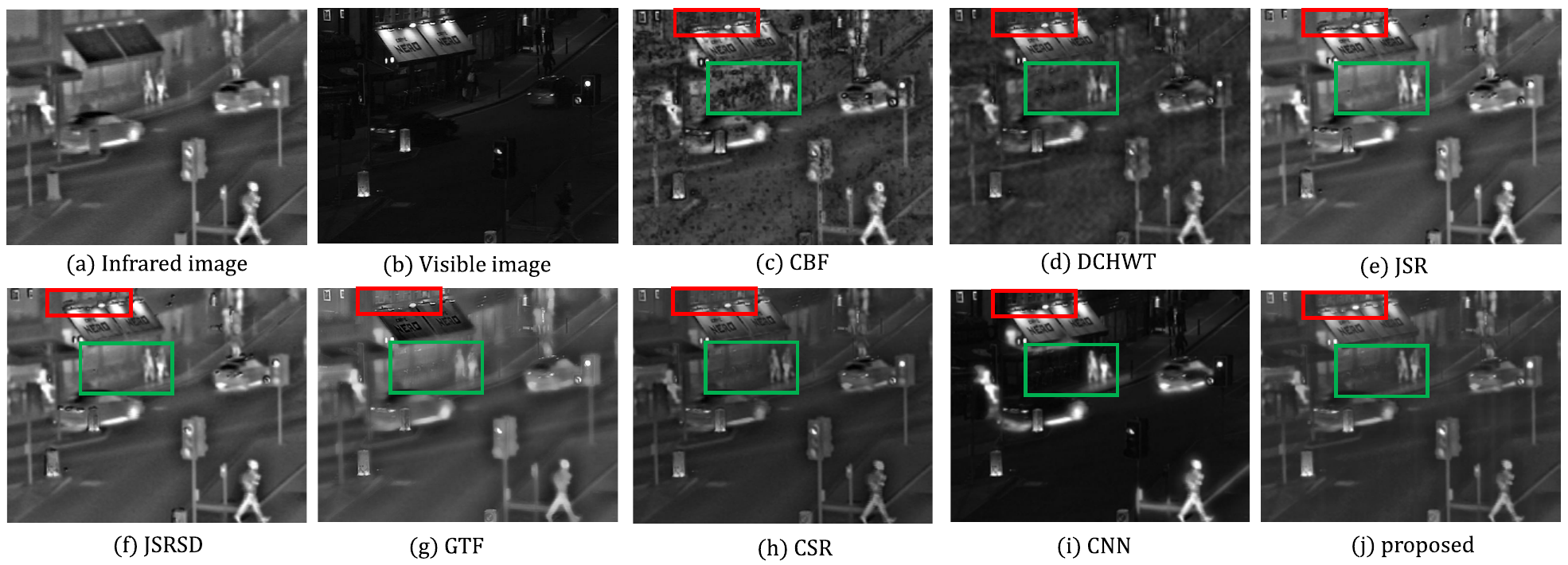}
\caption{Experiment on ``street'' images. (a) Infrared image; (b) Visible image; (c) CBF; (d) DCHWT; (e) JSR; (f) JSRSD. (g) GTF; (i) CSR; (h) CNN; (j) proposed method.}
\label{fig:street}
\end{figure*}

\begin{figure*}[!ht]
\centering
\includegraphics[width=\linewidth]{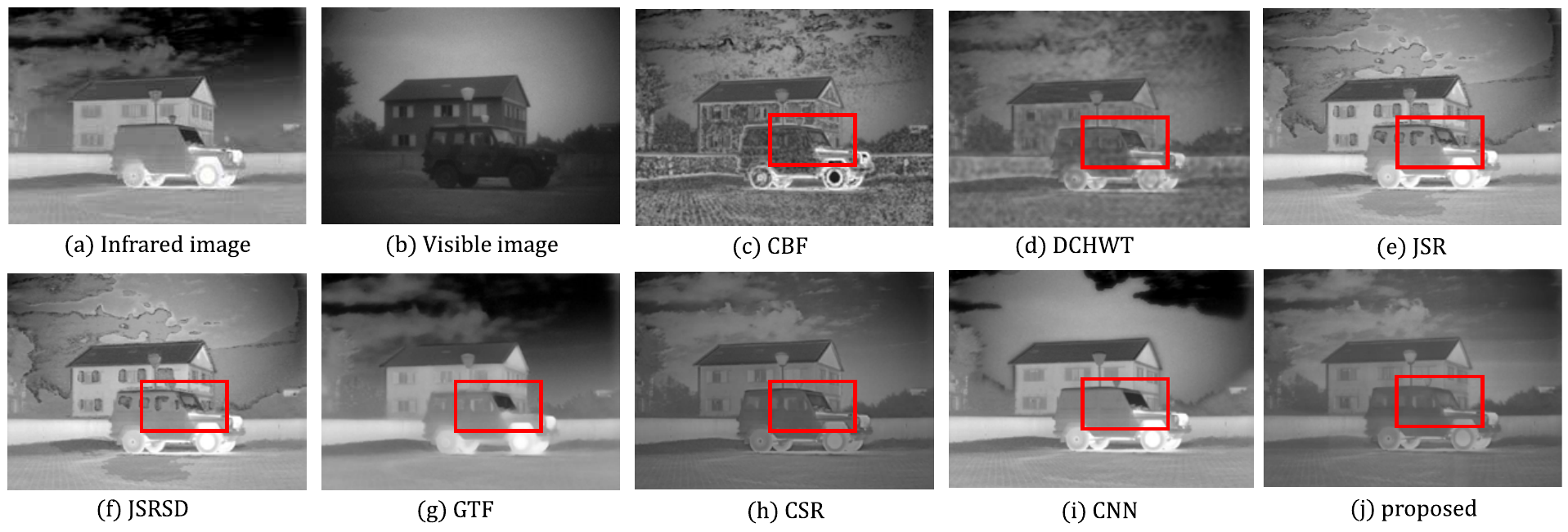}
\caption{Experiment on ``car'' images. (a) Infrared image; (b) Visible image; (c) CBF; (d) DCHWT; (e) JSR; (f) JSRSD. (g) GTF; (i) CSR; (h) CNN; (j) proposed method.}
\label{fig:car}
\end{figure*}

Then several novel and classical fusion methods are compared with the proposed method, including: cross bilateral filter fusion method(CBF)\cite{cbf2013}, discrete cosine harmonic wavelet transform fusion method(DCHWT)\cite{dchwt2012}, the joint-sparse representation model(JSR)\cite{jsr2013}, the JSR model with salient detection fusion method(JSRSD)\cite{jsrsd2017}, the gradient transfer fusion method(GTF)\cite{gtf2016}, convolutional sparse representation(CSR)\cite{csr2016} and deep convolutional neural network-based method(CNN)\cite{cnn2017}.

In our experiments, the parameter $\lambda$ in LatLRR is 0.8. And the weighting value for fusing the low-rank parts is 0.5. The evaluation methods for fusion performance will be introduced in the next subsections.

All the experiments are implemented in MATLAB R2016a on 3.2 GHz Intel(R) Core(TM) CPU with 12 GB RAM.

\subsection{Subjective Evaluation}
The fusion results are obtained by five compared methods and proposed method are shown in Fig.\ref{fig:street} - \ref{fig:car}, we just choose two pairs of infrared and visible images(``street'' and ``car'') to assess the fusion performance by subjective evaluation.

In Fig.\ref{fig:street} and Fig.\ref{fig:car}, Fig.\ref{fig:street}(c-j) and Fig.\ref{fig:car}(c-j) are the fused images which are obtained by the compared fusion methods and the proposed fusion method. As we can see from the fused images in Fig.\ref{fig:street}, the fused image which is obtained by proposed method can preserve more windows and chair detail information in red and green boxes, respectively. In Fig.\ref{fig:car}, the fused image which is obtained by proposed method also contain more glass detail information in red box.

However, the fused images which are obtained by CBF and DCHWT have more artifacts information and the salient features are not clear. For the fused images which are obtained by JSR, JSRSD, GTF, CSR and CNN contain many ringing artifacts around the salient features and the detail information are not clear either. In contrast, the fused images which are obtained by proposed fusion method contain more salient features and preserve more detail information. Compared with above fusion methods, the fused images obtained by proposed fusion method are more natural for human perception and the proposed method has better subjective fusion performance.

\begin{table*}[ht]
\scriptsize
\centering
\caption{\label{tab:table1}The average values of the compared methods and the proposed method for 21 pairs source images.}
\resizebox{\linewidth}{!}{
\begin{tabular}{|c|c|c|c|c|c|c|c|c|}
\hline
\emph{Methods} & CBF\cite{cbf2013} & DCHWT\cite{dchwt2012} & JSR\cite{jsr2013} & JSRSD\cite{jsrsd2017} & GTF\cite{gtf2016} &CSR\cite{csr2016} &CNN\cite{cnn2017} &Proposed \\
\hline
Qabf\cite{Qabf}  &0.43961	&0.46592	&0.32306	&0.32281	&0.41037	&\textbf{0.53485}	&0.28789 	&0.41277\\
\hline
SCD\cite{scd2015}  &1.38963 	&1.60993 	&1.59136 	&1.59124 	&1.00488 	&1.64823	&1.48060 	&\textbf{1.70699}\\
\hline
$SSIM_a$  &0.59957 	&0.73132 	&0.54073 	&0.54127 	&0.70016 	&0.7533	&0.71109 	&\textbf{0.76486}\\
\hline
$N_{abf}$\cite{dchwt2012}  &0.31727 	&0.12295 	&0.34712 	&0.34657 	&0.07951 	&0.01958	&0.02324 	&\textbf{0.01596}\\
\hline
\end{tabular}}
\end{table*}

\subsection{Objective Evaluation}
For the purpose of quantitative comparison between the proposed method and other fusion methods, four quality metrics are utilized. These are: Qabf\cite{Qabf}, $N_{abf}$\cite{dchwt2012} which denotes the rate of noise or artifacts added in the fused image due to fusion process, the sum of the correlations of differences(SCD)\cite{scd2015}, and modified structural similarity($SSIM_a$).

In our experiment, the $SSIM_a$ is calculated by Eq.(\ref{equ:ssim}),
\begin{eqnarray}\label{equ:ssim}
  	SSIM_{a}(F)=(SSIM(F,I_1)+SSIM(F,I_2))\times0.5
\end{eqnarray}

\noindent where $SSIM(\cdot)$ represents the structural similarity operation, $F$ is fused image. And $I_1$, $I_2$ are source images. The value of $SSIM_a$ denotes the ability of structural preservation.

The fusion performance of fused image is better with the increasing numerical index of Qabf, SCD and $SSIM_a$. On the contrary, the fusion performance is better when the value of $N_{abf}$ is small.

The average values of $Qabf$, SCD, $SSIM_a$ and $N_{abf}$ for 21 pairs of source images are shown in Table \ref{tab:table1}. 

In Table \ref{tab:table1}, the best values of quality metrics are indicated in bold. As we can see, the proposed method has the best values in SCD, $SSIM_a$ and $N_{abf}$, respectively. 

These values indicate that the fused images obtained by the proposed method are more natural and contain less artificial information. From objective evaluation, our fusion method has better fusion performance than those compared methods.

\section{Conclusions}

In this paper, we propose a simple yet effective infrared and visible image fusion method based on latent low-rank representation. Firstly, the source images are decomposed into low-rank parts and salient parts which contain global structure and local structure information, respectively. Then the fused low-rank part and salient part are obtained by weighted-average strategy, and we choose different weight value for these two parts. Finally, the fused image is reconstructed by adding the fused low-rank part and the fused salient part. We use both subjective and objective methods to evaluate the proposed method, the experimental results show that the proposed method exhibits better performance than other compared methods.

\bibliographystyle{ieee}
\bibliography{ivf_bib}

\begin{thebibliography}{10}\itemsep=-1pt

\bibitem{scd2015}
V.~Aslantas and E.~Bendes.
\newblock A new image quality metric for image fusion: the sum of the
  correlations of differences.
\newblock {\em Aeu-international Journal of electronics and communications},
  69(12):1890--1896, 2015.

\bibitem{twoscale2016}
D.~P. Bavirisetti and R.~Dhuli.
\newblock Two-scale image fusion of visible and infrared images using saliency
  detection.
\newblock {\em Infrared Physics \& Technology}, 76:52--64, 2016.

\bibitem{DWT2005}
A.~Ben~Hamza, Y.~He, H.~Krim, and A.~Willsky.
\newblock A multiscale approach to pixel-level image fusion.
\newblock {\em Integrated Computer-Aided Engineering}, 12(2):135--146, 2005.

\bibitem{gao2017}
R.~Gao, S.~A. Vorobyov, and H.~Zhao.
\newblock Image fusion with cosparse analysis operator.
\newblock {\em IEEE Signal Processing Letters}, 24(7):943--947, 2017.

\bibitem{dchwt2012}
B.~S. Kumar.
\newblock Multifocus and multispectral image fusion based on pixel significance
  using discrete cosine harmonic wavelet transform.
\newblock {\em Signal, Image and Video Processing}, 7(6):1125--1143, 2013.

\bibitem{cbf2013}
B.~S. Kumar.
\newblock Image fusion based on pixel significance using cross bilateral
  filter.
\newblock {\em Signal, image and video processing}, 9(5):1193--1204, 2015.

\bibitem{GithubHuiLi}
H.~Li.
\newblock Code: Infrared and visible image fusion using latent low-rank
  representation, 2018.
\newblock https://github.com/hli1221/imagefusion\_Infrared\_visible\_latlrr.

\bibitem{lrr2017}
H.~Li and X.-J. Wu.
\newblock Multi-focus image fusion using dictionary learning and low-rank
  representation.
\newblock In {\em International Conference on Image and Graphics}, pages
  675--686. Springer, 2017.

\bibitem{li2018densefuse}
H.~Li and X.-J. Wu.
\newblock Densefuse: A fusion approach to infrared and visible images.
\newblock {\em IEEE Transactions on Image Processing}, 28(5):2614--2623, 2018.

\bibitem{li2018infrared}
H.~Li, X.-J. Wu, and J.~Kittler.
\newblock Infrared and visible image fusion using a deep learning framework.
\newblock In {\em 2018 24th International Conference on Pattern Recognition
  (ICPR)}, pages 2705--2710. IEEE, 2018.

\bibitem{Shutao2017}
S.~Li, X.~Kang, L.~Fang, J.~Hu, and H.~Yin.
\newblock Pixel-level image fusion: A survey of the state of the art.
\newblock {\em Information Fusion}, 33:100--112, 2017.

\bibitem{jsrsd2017}
C.~Liu, Y.~Qi, and W.~Ding.
\newblock Infrared and visible image fusion method based on saliency detection
  in sparse domain.
\newblock {\em Infrared Physics \& Technology}, 83:94--102, 2017.

\bibitem{Lrr2010}
G.~Liu, Z.~Lin, and Y.~Yu.
\newblock Robust subspace segmentation by low-rank representation.
\newblock In {\em ICML}, volume~1, page~8, 2010.

\bibitem{latentLrr2011}
G.~Liu and S.~Yan.
\newblock Latent low-rank representation for subspace segmentation and feature
  extraction.
\newblock In {\em 2011 International Conference on Computer Vision}, pages
  1615--1622. IEEE, 2011.

\bibitem{cnn2017}
Y.~Liu, X.~Chen, H.~Peng, and Z.~Wang.
\newblock Multi-focus image fusion with a deep convolutional neural network.
\newblock {\em Information Fusion}, 36:191--207, 2017.

\bibitem{csr2016}
Y.~Liu, X.~Chen, R.~K. Ward, and Z.~J. Wang.
\newblock Image fusion with convolutional sparse representation.
\newblock {\em IEEE signal processing letters}, 23(12):1882--1886, 2016.

\bibitem{ivsr2014}
X.~Lu, B.~Zhang, Y.~Zhao, H.~Liu, and H.~Pei.
\newblock The infrared and visible image fusion algorithm based on target
  separation and sparse representation.
\newblock {\em Infrared Physics \& Technology}, 67:397--407, 2014.

\bibitem{luo2016novel}
X.~Luo, Z.~Zhang, and X.~Wu.
\newblock A novel algorithm of remote sensing image fusion based on
  shift-invariant shearlet transform and regional selection.
\newblock {\em AEU-International Journal of Electronics and Communications},
  70(2):186--197, 2016.

\bibitem{nonshearlet2017}
X.~Luo, Z.~Zhang, B.~Zhang, and X.~Wu.
\newblock Image fusion with contextual statistical similarity and nonsubsampled
  shearlet transform.
\newblock {\em IEEE Sensors Journal}, 17(6):1760--1771, 2016.

\bibitem{Luo2017JVCIR}
X.~Luo, Z.~Zhang, C.~Zhang, and X.~Wu.
\newblock Multi-focus image fusion using hosvd and edge intensity.
\newblock {\em Journal of Visual Communication and Image Representation},
  45:46--61, 2017.

\bibitem{gtf2016}
J.~Ma, C.~Chen, C.~Li, and J.~Huang.
\newblock Infrared and visible image fusion via gradient transfer and total
  variation minimization.
\newblock {\em Information Fusion}, 31:100--109, 2016.

\bibitem{ma2019fusiongan}
J.~Ma, W.~Yu, P.~Liang, C.~Li, and J.~Jiang.
\newblock Fusiongan: A generative adversarial network for infrared and visible
  image fusion.
\newblock {\em Information Fusion}, 48:11--26, 2019.

\bibitem{ivwls2017}
J.~Ma, Z.~Zhou, B.~Wang, and H.~Zong.
\newblock Infrared and visible image fusion based on visual saliency map and
  weighted least square optimization.
\newblock {\em Infrared Physics \& Technology}, 82:8--17, 2017.

\bibitem{Manchanda2018}
M.~Manchanda and R.~Sharma.
\newblock An improved multimodal medical image fusion algorithm based on fuzzy
  transform.
\newblock {\em Journal of Visual Communication and Image Representation},
  51:76--94, 2018.

\bibitem{sun2011quantum}
J.~Sun, W.~Fang, X.~Wu, and W.~Xu.
\newblock Quantum-behaved particle swarm optimization: principles and
  applications, 2011.

\bibitem{TNO2014}
A.~Toet.
\newblock Tno image fusion dataset, 2014.
\newblock https://figshare.com/articles/TN\_Image\_Fusion\_Dataset/1008029.

\bibitem{shearlet2014}
L.~Wang, B.~Li, and L.-F. Tian.
\newblock Eggdd: An explicit dependency model for multi-modal medical image
  fusion in shift-invariant shearlet transform domain.
\newblock {\em Information Fusion}, 19:29--37, 2014.

\bibitem{wang2003initial}
M.~Wang, S.-T. Wang, and X.-J. Wu.
\newblock Initial results on fuzzy morphological associative memories.
\newblock {\em Acta Electronica Sinica}, 31(5):690--693, 2003.

\bibitem{Qabf}
C.~Xydeas, , and V.~Petrovic.
\newblock Objective image fusion performance measure.
\newblock {\em Electronics letters}, 36(4):308--309, 2000.

\bibitem{contourlet2010}
S.~Yang, M.~Wang, L.~Jiao, R.~Wu, and Z.~Wang.
\newblock Image fusion based on a new contourlet packet.
\newblock {\em Information Fusion}, 11(2):78--84, 2010.

\bibitem{ivsidt2016}
M.~Yin, P.~Duan, W.~Liu, and X.~Liang.
\newblock A novel infrared and visible image fusion algorithm based on
  shift-invariant dual-tree complex shearlet transform and sparse
  representation.
\newblock {\em Neurocomputing}, 226:182--191, 2017.

\bibitem{jsr2013}
Q.~Zhang, Y.~Fu, H.~Li, and J.~Zou.
\newblock Dictionary learning method for joint sparse representation-based
  image fusion.
\newblock {\em Optical Engineering}, 52(5):057006, 2013.

\bibitem{morphological2017}
Y.~Zhang, X.~Bai, and T.~Wang.
\newblock Boundary finding based multi-focus image fusion through multi-scale
  morphological focus-measure.
\newblock {\em Information fusion}, 35:81--101, 2017.

\bibitem{zheng2006nearest}
Y.-J. Zheng, J.-Y. Yang, J.~Yang, X.-J. Wu, and Z.~Jin.
\newblock Nearest neighbour line nonparametric discriminant analysis for
  feature extraction.
\newblock {\em Electronics Letters}, 42(12):679--680, 2006.

\bibitem{medicalclassified2017}
J.-j. Zong and T.-s. Qiu.
\newblock Medical image fusion based on sparse representation of classified
  image patches.
\newblock {\em Biomedical Signal Processing and Control}, 34:195--205, 2017.

\end{thebibliography}

\end{document}